\documentclass{article}
\usepackage{hyperref}
\begin{document}

\title{PHILOSOPHY IN THE FACE OF ARTIFICIAL INTELLIGENCE\thanks{An edited
    version of this article was published in the magazine {\em Prospect}
    under the title
    \href{http://www.prospectmagazine.co.uk/science-and-technology/artificial-intelligence-wheres-the-philosophical-scrutiny}{``Artificial
      intelligence:~where's the philosophical scrutiny?''}~on May 4,
    2016~\cite{Conitzer16:Artificial}.  I thank Tom Dietterich and Walter
    Sinnott-Armstrong for helpful feedback on earlier versions of this
    article.}}  \author{Vincent Conitzer\\Duke University\\Durham, NC, USA}
\date{}
\maketitle

The idea of Artificial Intelligence has captured our collective imagination
for many decades.  Can behavior that we think of as intelligent be
replicated on a machine?  If so, what consequences could this have for
society?  And what does it tell us about ourselves as human beings?
Besides being a long-running topic of philosophical reflection and science
fiction, AI is also a well-established scientific research area.  Many
universities have AI research labs, usually housed in computer science
departments.  The feats accomplished in such research have generally been
far more modest than those imagined in the movies.  But over time, the gap
between reality and fiction has been closing.  For example, self-driving
cars are now a reality.  And the world outside academia has taken notice.
The commercial opportunities are endless and technology companies are in
fierce competition over the top AI talent.  Meanwhile, there is a growing
popular worry about where this is all headed.

Most of the technical progress on AI is reported at scientific conferences
on the subject.  These conferences have been running for decades and are
attended by a steady community of devoted researchers.  But in recent years
they have also started to attract a broader mix of participants.  At the
2016 conference in Phoenix, one speaker was more controversial than any
other in recent memory: Nick Bostrom.  While the audience consisted mostly
of computer scientists, Bostrom is a philosopher who directs the
\href{https://www.fhi.ox.ac.uk/}{Future of Humanity Institute} at Oxford.
He recently made waves with his book
\href{https://global.oup.com/academic/product/superintelligence-9780199678112}{{\em
    Superintelligence}}~\cite{Bostrom14:Superintelligence}.  In it, he
contemplates the problem that we may soon build AI that broadly exceeds
human capabilities, and considers what steps we can take now to ensure that
the result will be in our best interest.  A key concern is that of an
``intelligence explosion'': if we are intelligent enough to build a machine
more intelligent than ourselves, then, so the thinking goes, surely that
machine in turn would be capable of building something even more
intelligent, and so on.  The phrase ``technological singularity'' is
sometimes also used to describe such runaway intelligence.  Will humanity
be left in the dust?  Will we be wiped out?  Since the appearance of
Bostrom's book, public figures including Elon Musk, Stephen Hawking, and
Bill Gates have warned of the risks of superintelligent AI.  Musk even
donated \$10M to the Boston-based \href{http://futureoflife.org/}{Future of
  Life Institute}, establishing a grants program to ensure that AI remains
beneficial.  The topic has remained in the news, with, for example,
recently United Nations Chief Information Technology Officer Atefeh Riazi
joining the chorus emphasizing the risks of AI.

These concerns have mostly been raised by people outside the core AI
research community, which has not been very vocal in this debate.  Some in
the community cautiously agree with some of the points; others dismiss
them.  As an AI researcher myself, after Bostrom's talk I saw a number of
people express their displeasure on social media, saying that giving him
such a forum gives him credibility that he does not deserve.  Others
emphasized open-mindedness, but (as far as I saw) fell short of endorsing
his ideas.  But I assume that most in the AI community shrugged and
continued with their research as usual.  Why?  Do AI researchers just not
care about the future of humanity?

I think the real answer requires some familiarity with the history of AI
research, which took off in the fifties.  Early research showed that
computers could do things that few at the time had expected, leading to
excitement, optimism, and promises of the moon.  But limitations of this
early work soon became apparent.  Approaches that produced impressive
results on small, toy examples simply would not scale to real-world
problems.  Also, the real world is messy and ambiguous, and AI researchers
struggle to this day with making their programs robust enough to handle
this.  This led to what was called an ``AI winter'': AI got a bad
reputation in the academic community and funding was reduced.  In fact,
this cycle repeated itself.  AI researchers yearned for their work to be
scientifically rigorous and respected, and learned to be careful.  Some
sought to dissociate themselves from the term ``AI'' altogether and instead
associated with more narrowly defined technical problems.  For example,
many researchers in the machine learning community -- which focuses on
having computers learn automatically from data how to make predictions and
decisions -- no longer wanted to be considered ``AI'' researchers.  Even
most of the researchers that did stick with the term started focusing on
narrower problems, not only because of perception issues but also for
technical reasons: these problems seemed to be important roadblocks for AI
but were not easy to solve.  Also, progress on their solution often led to
direct beneficial impact on society.  For example, part of the community
has focused on automated planning and scheduling systems, which have been
used in a variety of applications, such as scheduling the observations of
the Hubble Space Telescope.

The AI community has also mostly avoided the philosophical issues.  An
introductory AI course will typically spend a little time on basic
philosophical questions, such as those raised by Searle's ``Chinese Room
Argument''~\cite{Searle80:Minds}.  In this argument, someone who does not
know Chinese at all sits in a room and has an incredibly detailed
step-by-step manual -- read, a computer program -- for how to respond to
Chinese characters slipped under the door, by drawing other characters and
slipping them back out.  The manual is so good that from the outside it
appears that there is someone inside who speaks Chinese, no matter how
sophisticated the questions posed.  Now, we can ask whether there is any
real understanding of Chinese in the room.  At first, it may appear that
there is not.  But if not, then how could a computer, which operates
similarly, ever have any real understanding?

While some AI professors enjoy posing such conundrums in, for example, the
introductory lecture, after that the typical AI course -- my own included
-- will quickly move on to teaching technical material that can be used to
create programs that do something interesting, like playing the game of
Connect Four.  After all, the course is generally taught in a computer
science department, not a philosophy department.  Similarly, very little of
the research presented at any major AI conference is philosophical in
nature.  Most of it comes in the form of technical progress -- a better
algorithm for solving an established problem, say.  This is where AI
researchers believe they can make useful progress and win respect in the
eyes of their scientific peers, whether they feel the philosophical
problems are important or not.

All this explains some of the reluctance of the AI community to engage with
the superintelligence debate.  It has fought very hard to establish itself
as a respected scientific discipline, overcoming outside bias and its own
careless early claims.  The mindset is that anything perceived as
unsubstantiated hype, or as being outside the realm of science, is to be
avoided at all costs.  Tellingly, in a panel after Bostrom's talk, Oren
Etzioni, director of the \href{http://allenai.org/}{Allen Institute for
  Artificial Intelligence}, drew supportive laughs from the crowd when he
pointed out that Bostrom's talk was blissfully devoid of any data -- even
though Etzioni was quick to acknowledge that this was inherent in the
problem.  Tom Dietterich, a computer science professor at Oregon State and
President of the \href{http://www.aaai.org/home.html}{Association for the
  Advancement of Artificial Intelligence}, expressed skepticism that an
intelligence explosion of the kind Bostrom describes would happen, and
asked what experiments we could run to test this hypothesis.  The AI
community generally eschews speculation about the deep future and is more
comfortable engaging with important problems that are more concrete and
tangible at this point, such as autonomous weapons -- weapon systems that
can act without human intervention -- or the unemployment caused by AI
replacing human workers.  The latter was, in fact, the topic of the panel.

Another issue is that AI researchers, perhaps unlike the general public,
generally feel that there are still quite a few needed components missing
before something like the superintelligent AI of Bostrom's book could
possibly emerge.  Many of the problems that were once thought to be great
benchmark problems for AI -- say, beating human champions at chess -- ended
up being solved using special-purpose techniques that, while impressive,
could not immediately be used to solve many other problems in AI,
suggesting that the ``hard problems'' of AI lay elsewhere.  (This has also
led AI researchers to lament that ``once we solve something, it's not
considered AI anymore.'')  So while recent breakthroughs, such as
\href{https://deepmind.com/}{Google DeepMind}'s AI learning to play old
Atari games surprisingly well, may raise concern in the general public,
perhaps AI researchers have become accustomed to the idea that this just
means the hard problems must lie elsewhere.  That being said, these results
are certainly impressive to the AI community as well, not least because
this time there are common techniques -- now generally referred to as
``deep learning'' -- underlying not only the Atari results, but also
surprising progress in speech and image recognition.  (Consider the
problems that Apple needs to solve to get Siri to understand what you said,
or that Facebook needs to solve to automatically recognize faces in the
pictures you upload.)  Researchers had previously attacked these problems
with separate special-purpose techniques.  And Google DeepMind's AlphaGo
program, which recently defeated Lee Sedol, possibly the best human player,
in the game of Go, also has deep learning at its core.  The techniques used
for chess had been largely ineffective on Go.

It is worth noting that the line of research that led to the deep learning
results had been largely dismissed by most AI and machine learning
researchers, before the few that tenaciously stuck with it started
producing impressive results.  So our predictions about how AI will
progress can be far off even in the very short term.  Accurately predicting
all the way to, say, the end of the century seems humanly impossible.  If
we go equally far into the past, we end up at a time before even Alan
Turing's 1936 paper that laid the theoretical foundation for computer
science~\cite{Turing36:On}.  This, too, makes it difficult for mainstream
AI researchers to connect with those raising concerns about the future.
Some disaster scenarios, such as those related to asteroid strikes or
global warming, allow for reasonable predictions over such timescales, so
it is natural to want the same for AI.  But AI researchers and computer
scientists in general tend to reason over much shorter timescales, which is
already challenging given the pace of progress.

As one of the recipients of a Musk-funded Future of Life Institute grant, I
participated on a keeping-AI-beneficial panel in a workshop at the
conference in Phoenix.  The panel was moderated by Max Tegmark, one of the
founders of the Future of Life Institute and a physics professor at MIT --
again, an outsider to the AI community.  Besides relatively more accessible
questions about autonomous weapons and technological unemployment, Tegmark
also asked the panel some philosophical questions.  All other things being
equal, would you want your artificially intelligent virtual assistant
(imagine an enormously improved Siri) to be conscious?  Would you want it
to be able to feel pain?  The first question had no takers; some in
attendance argued that pain could be beneficial from the perspective of the
AI learning to avoid bad actions.  The substantial philosophical literature
on consciousness and qualia did not come up.  (In philosophy, the word
``qualia'' refers to subjective experiences, such as pain, and more
specifically to what it is {\em like} to have the experience.  A famous
example due to the philosopher Thomas Nagel is that presumably, there is
something it is like to be a bat, though we, as a species that does not use
echolocation, may never know exactly what this is like~\cite{Nagel74:What}.
Is there something it is like to be an AI virtual assistant?  A
self-driving car?)  Perhaps this was less due to unfamiliarity with such
concepts, and more due to discomfort with how to approach these questions.
Even philosophers have difficulty agreeing on the meaning of these terms,
and the literature ranges from the more scientifically oriented search for
the ``neural correlates of consciousness'' (roughly: what is going on in
the brain when conscious experience takes place) all the way to more
esoteric studies of the subjective: how is it that {\em my} subjective
experiences so appear so vividly {\em present}, while yours do not?  Well,
surely your experiences appear similarly {\em somewhere else}.  Where?  In
your brain, as opposed to mine?  But when we inspect a brain, we do not
find any qualia, just neurons.  (If all this seems hopelessly obscure to
you, you are not alone -- but if you are intrigued, see, for example,
Caspar Hare's \href{http://press.princeton.edu/titles/8921.html}{{\em On
    Myself, and Other, Less Important Subjects}}~\cite{Hare09:On} or
J.J.~Valberg's \href{http://press.princeton.edu/titles/8416.html}{{\em
    Dream, Death, and the Self}}~\cite{Valberg07:Dream}.)  The state of our
understanding makes it difficult even to agree on what exactly Tegmark's
questions mean -- is objectively assessing whether an AI virtual assistant
has subjective experiences a contradiction in terms? -- let alone give
actionable advice to AI practitioners.  I believe philosophers do make
progress on these issues, but it is slow and hard-won.  When discussing
what philosophers are to do, Bostrom in his book suggests to postpone work
on ``some of the eternal questions'' for a while, and instead to focus on
how best to make it through the transition to a world with superintelligent
AI.  But it is not entirely clear whether and how we can sidestep the
eternal questions in this endeavor, even if we accept the premise that such
a transition will take place.  (Of course, philosophers do not necessarily
accept the premise either.)

So, generally, AI researchers prefer to avoid these questions and return to
making progress on more tractable problems.  Many of us are driven to make
the world a better place -- by reducing the number of deaths from
automobile accidents, increasing access to education, improving
sustainability and healthcare, preventing terrorist attacks, etc.~-- and
are a bit frustrated to see every other article on AI in the news
accompanied by an image from {\em The Terminator}.  Meanwhile, genuine
concerns are developing outside the AI community.  While the AI conference
in Phoenix was already underway, there was a call at a meeting of the
\href{http://www.aaas.org/}{American Association for the Advancement of
  Science} to devote 10\% of the AI research budget to the study of the
societal impact of AI.  Now, in a climate where funding is already tight,
diverting some of it may not make AI researchers look more kindly on the
people raising these concerns.  But the AI community should take part in
the debate on societal impact, because without it the debate will still
take place and be less informed.  Fortunately, members of the community are
increasingly taking an interest in short-to-medium-term policy questions,
including
\href{http://futureoflife.org/open-letter-autonomous-weapons/}{calling for
  a ban on autonomous weapons}.  Unfortunately, we have yet to figure out
how to rigorously and productively engage with the more nebulous long-term
philosophical issues.  One area where some immediate traction seems
possible is the study of how (pre-superintelligence) AI can make ethical
decisions -- for example, when a self-driving car needs to make a decision
in a scenario that is likely to kill or injure someone.  In fact, automated
ethical decision making is the topic of a number of the Future of Life
Institute grants, including my own grant with Walter Sinnott-Armstrong, a
professor of practical ethics and philosophy at Duke University.  But at
this point it is not clear to AI researchers how to usefully address the
notion of superintelligence and the philosophical questions raised by it.

At the end of Bostrom's talk, Moshe Vardi, a computer science professor at
Rice University, suggested that this all was very much as if upon Watson
and Crick's discovery of the structure of DNA, the focus had immediately
been on all the ways in which it could be abused.  I think this is an
excellent point.  Progress in AI will unfold in unexpected ways and some of
the current concerns will turn out to be unfounded, especially among those
concerning the far off future.  But this argument cuts both ways; we can be
sure that there are risks that are not currently appreciated.  It is not
clear what exact course of action is called for, but those that know the
most about AI cannot be complacent.

\bibliographystyle{plain}

\end{document}